# Medium. Permeation: SARS-COV-2 Painting Creation by Generative Model

Yuan-Fu Yang

yfyangd@gmail.com

Iuan-Kai Fang

iuankai.fang@gmail.com

Min Sun

sunmin@ee.nthu.edu.tw

Su-Chu Hsu

suchu@mx.nthu.edu.tw

## Abstract

*Airborne particles are the medium for SARS-CoV-2 to invade the human body. Light also reflects through suspended particles in the air, allowing people to see a colorful world. Impressionism is the most prominent art school that explores the spectrum of color created through color reflection of light. We find similarities of color structure and color stacking in the Impressionist paintings and the illustrations of the novel coronavirus by artists around the world. With computerized data analysis through the main tones, the way of color layout, and the way of color stacking in the paintings of the Impressionists, we train computers to draw the novel coronavirus in an Impressionist style using a Generative Adversarial Network to create our artwork "Medium. Permeation". This artwork is composed of 196 randomly generated viral pictures arranged in a 14×14 matrix to form a large-scale painting.*

*In addition, we have developed an extended work: Gradual Change, which is presented as video art. We use Graph Neural Network to present 196 paintings of the new coronavirus to the audience one by one in a gradual manner. In front of LED TV screen, audience will find 196 virus paintings whose colors will change continuously. This large video painting symbolizes that worldwide 196 countries have been invaded by the epidemic, and every nation continuously pops up mutant viruses. The speed of vaccine development cannot keep up with the speed of virus mutation. This is also the first generative art in the world based on the common features and a metaphorical symbiosis between Impressionist art and the novel coronavirus. This work warns us of the unprecedented challenges posed by the SARS-CoV-2, implying that the world should not ignore the invisible enemy who uses air as a medium.*

*Keywords: SARS-CoV-2; Generative Art; Graph Neural Network*

## 1. Introduction

Since the advent of artificial intelligence, scientists have been exploring the ability of machines to generate human-level creative products such as poetry, stories, music, and paintings. This ability is proving that artificial intelligence algorithms are the foundation of human intelligence. In the visual arts, several systems for automatic creation by machines have been proposed, not only in the fields of artificial intelligence and computational creativity, but also in the fields of computer graphics and machine learning. Our work is a generative art, using Generative Adversarial Network to train a computer to draw the novel coronavirus in an Impressionist style.

Virus particles smaller than 5 μm will be temporarily suspended in the air, and the virus will enter the human body through a scattering path. On January 21, 2020, CDC (Centers of Disease Control and Prevention) illustrators Alissa Eckert and Dan Higgins were asked to illustrate the novel coronavirus for use in press. SARS-CoV-2 has since given birth to colorful spherical shapes. When light encounters particles in the air, it will produce a scattering state, and the air is full of oxygen and nitrogen molecules, whose size is even shorter than the wavelength of short wavelengths, and the scattered light will be in the range of violet, blue and green. When the light in the blue wavelength band is scattered away, red-orange-yellow colors appear. Impressionism expressed the scientific significance of light in artistic creation. The light color reflected by the luminous flux through the medium in the air became the theoretical basis of Impressionist creation.

The purpose of this work is to study a computational creativity system that can be used for SARS-CoV-2 painting generation, which does not involve human artists in the creation process, but involves human creativity in the learning process. Therefore, we collected worldwide illustrators' specific imagination of SARS-CoV-2 under the microscope, and developed a Generative Adversarial Network to learn these illustrations. We tune various parameters during training. Through this deep learning work, the machine imitated over 300 viral illustrations we collected around the world into the painting styles of Impressionist painters Dou Jia, Monet, and Renoir. In this style, we try to express the scattering path of SARS-CoV-2 through the air is like the scattering phenomenon of light molecules caused by the air.

We propose a deep learning approach for generator image super-resolution. Our method directly learns the end-to-end mapping between low and high-resolution images. The map is represented as a deep convolutional neural network that takes a low-resolution image as input and outputs a high-resolution image. In addition, we developed an extension work: incremental changes. We used Graph Neural Network to make 196 virus images are collaged into a moving picture which is presented by a random gradient system, like 196 brightly blooming flowers, constantly changing self-colors.

We hope reminding people's heart searching, meanwhile, this work also includes artistic aesthetics. People are afraid of viruses that cannot be exterminated, but we hope that the creation itself can free the viewers from fear, so that they can get close to our artwork. It will change in real time and have aesthetic art allowing viewers to stop a while for gazing it, and peacefully think about the connotation we want to convey, so as to achieve the original intention of this artwork.

## 2. Related Work

After SARS-CoV-2 virus outbreaking, more and more artists use SARS-CoV-2 as their creative motivation. When we create this work, we try to understand the impact of different artists or art teams their purpose of the creation form and content of SARS-CoV-2 then try to choose a kind that can fully express the impact of the virus's constant mutation in the world, and we believe that this impact is comprehensive, so it is imperative for any individual to reflect on the problems faced by themselves and others. One of the concepts in our work is a metaphor that the natural environment treats human beings equally.

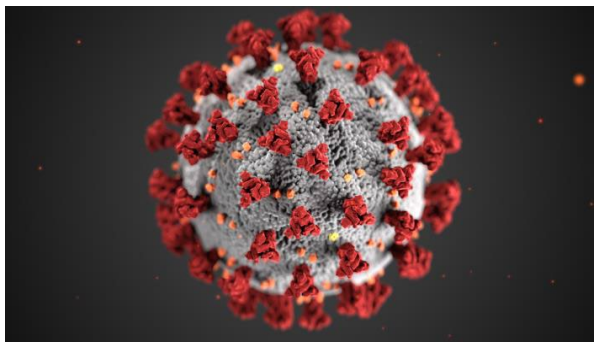

Figure 1: Impressions of SARS-CoV-2 by Alissa Eckert.

### 2.1. Coronavirus Arts

Since 2020, many artists have produced impressions of SARS-CoV-2 for government agencies and healthcare organizations to inform the general public of the virus' dangers and disseminate disease prevention measures. Alissa Eckert [1], a medical illustrator, was the first to assist the CDC to produce and publish a three-dimensional image of the coronavirus. The image was made using database information on proteins and their coordinates. The information was downloaded to a visualization software and using 3Ds Max and After Effect software, an image of the novel coronavirus was produced after 3D rendering of the proteins and adding shades and texture (as shown in Figure 1). She described her creative conceptualization of the virus' color and texture as the following: "I was thinking about making a velvety texture on the proteins, and something that looked like you could touch it and feel it. And I also wanted it to be solid, a bit rocky, something found in nature. Because if you relate it to something that exists, it's going to be more believable." [1].

Another molecular artist David Goodsell [2] used Fortran to create a customized computer colorization algorithm. He accessed the constitutive parameters for protein structure formation and converted the results into the geometric architecture of the novel coronavirus, then adding the final touches with watercolor. Goodsell made many freely downloadable images available on the RCSB Protein Data Bank website for promotional medical education purposes [2]. Figure 2 uses a minuscule droplet to demonstrate a cross-section view of those droplets that are thought to be disseminating SARS-CoV-2 viruses. The virus is colored in magenta. The droplet is filled with molecules commonly found in the respiratory tract, including green mucoproteins , blue pulmonary surfactants and lipids, as well as maroon antibodies.

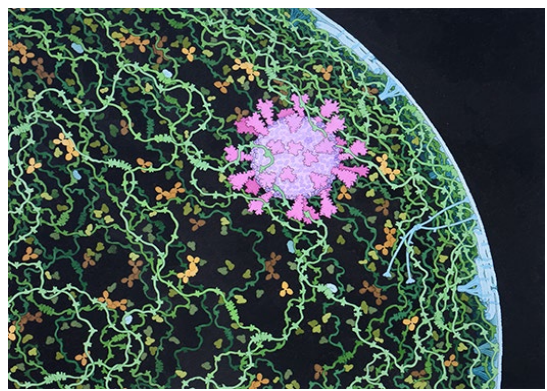

Figure 2: Illustration of SARS-CoV-2 by David Goodsell.

The 3D Visualization Aesthetics Lab of the University of New South Wales (UNSW) created a video by scientifically accurate simulation, which shows soap acting on contaminated skin covered with tiny coronavirus particles [3]. Their 3D visualization techniques successfully synthesized information on the particulate structure and composition of a SARS-CoV-2 virus into scientific 3D data through computational graphics. The complete virus particle is created using various spherical structures of proteins and lipids as the basic filler geometrical shapes processed by 3D graphics tools (as shown in Figure 3). In addition to encouraging everyone to wash their hands frequently with soap to prevent the spread of the epidemic, this film also conveys the collaboration of science and art. They applied data technology to the three-dimensional composition of virus particles, which inspired us to use data technology as the basis for drawing viruses, so we conceived of converting hundreds of collected virus illustrations into numerical values and trained the computer recognizing the outline of virus and its color.

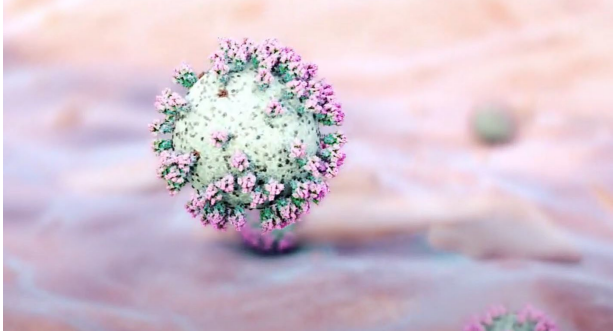

Figure 3: A 3D-visualisation of soap destroying the coronavirus by UNSW.

## 2.2. Color & Impressionism

The impressionists' method to render color and light is a major innovation in the history of art. Impressionist painters emphasize the delineation of light, highlighting colors as the main constitutive elements of a scene, at the same time fading out the sense of mass with the objects. They do not depend on contrast and linearly formed spatial distance. Impressionist painters limit the colors on their palettes, and instead adhere to light reflexivity principles. They preferred optical mixing instead of mixing pigments on their palettes. They also lace together cold and warm colors to form spatial effects [4].

Impressionist painters use red, yellow, and blue as main colors, then use overlapping and complementary colors to create new hues. Contrasts between red and green, yellow and violet, blue and orange create vibrant visual contrasts and a new sense of harmony. Impressionism revolutionized the classicism tradition and focused more on plein-air painting as opposed to studio work. Impressionists align with the École de Barbizon, but they used much more pure colors and replaced the solemn brown typical of the classicist tradition.

After the 18th century, the rise of Rationalism and the coming of the first Industrial Revolution brought considerable attention of the general public to technology. People were interested in science, and so were the impressionist painters. They revered the revolutionary products brought about by scientific development, and their research into painting techniques were also brimming with the scientific spirit, particularly their study of light and color. The colors of the Impressionism era challenged traditional and object-oriented art history. That was one of the most attractive chapters of modern visual culture [5].

## 2.3. Generative Arts

Philip Galanter [6] defines generative art as any artistic practice in which artists applied systems such as natural language patterns, computer programs, machines, or other program innovations, as these systems have to a certain degree activated its own autonomous contribution to or even completed the work of art. He gives three specific definitions for generative art: 1) the work of art must include known clusters of past and current generative art activity, 2) allow for forms of generative art yet to be discovered, 3) exist as a subset of all art, and allowing for contestation of the definition of art.

Our concept in this paper conforms to Philip Galanter's definition of generative art. We collected current clusters of SARS-CoV-2 illustrations from many artists, and we included past known clusters, namely the impressionist paintings. The two sets of data were then used as materials to train machines for recognition, and impressionist styles were imported to guide the rendering process of SARS-CoV-2. The aim is to create the conditions in which light particles reflected off the image must travel through air just as virus particles to achieve their purpose. But computer-generated graphics are not predictable: there are modes of expression unknown to us before the actual creative process, and we relegate the final visualized result to the computer for algorithmic generation.

We also studied Marc Lee's SARS-CoV-2 work Corona TV Bot [7]. He also used a certain algorithmic generative process, which creates a TV show by randomly capturing social media updates and has them appear simultaneously on one screen. Since the pandemic, Marc Lee would record six hours of social media posts at different times during the day and in the night every eight days to capture different pandemic related news from different parts of the world. It forms a history-based depository that brings together professional broadcasting videos across the globe, as well as any personal content tagged and published on Twitter and YouTube with hashtags #Coronavirus and #COVID-19 [8]. Apparently, Marc Lee collected world-wide text-based information as a data set as a randomly generated creative work. In comparison, our project collects images as our randomly generated data source.

Generative art is exactly a method that uses computer programming as a form of artistic creation. Lioret [9] showed how to use new quantum tools to achieve original generative creations, whether for images, 3D sculptures or animations. His lab used the famous Schrödinger equation to generate quantum animations [9]. He is very bullish on the potential of using quantum-generated adversarial network (QGAN) to create art works. Even if there are massive amounts of data that needs to be processed using this method, future quantum computers loaded with autonomously generated adversarial procedures will have even more efficiency for training the computer to produce frameworks closely aligned to the creator's intent.

## 2.4. Generative Adversarial Network

In the computational creativity literature, different algorithms have been proposed focused on investigating various and effective ways of exploring the creative space. Several approaches have used an evolutionary process in which the algorithm iterates by generating candidates, evaluating them using a fitness function, and then modifying them to improve the fitness score for the next iteration. Typically, this process is done within a genetic algorithm framework. As pointed out by DiPaola and Gabora [10], the challenge of any algorithm centers on "how to write a logical fitness function that has an aesthetic sense". Some early systems utilized a human in the loop to guide the art generating process [11]. In these interactive systems, the computer explores the creative space, and the human plays the role of the observer whose feedback is essential in driving the process.

The most famous interactive system in recent years is Generative Adversarial Network (GANs) [12]. GANs has two sub networks, a generator and a discriminator (as shown in Figure 4). The discriminator has access to a set of training images. The discriminator tries to discriminate between "real" images from the training set and "fake" images generated by the generator. The generator tries to generate images similar to the training set without seeing these images. The generator starts by generating random images and receives a signal from the discriminator whether the discriminator finds them real or fake. At equilibrium the discriminator should not be able to tell the difference between the images generated by the generator and the actual images in the training set, hence the generator succeeds in generating images that come from the same distribution as the training set.

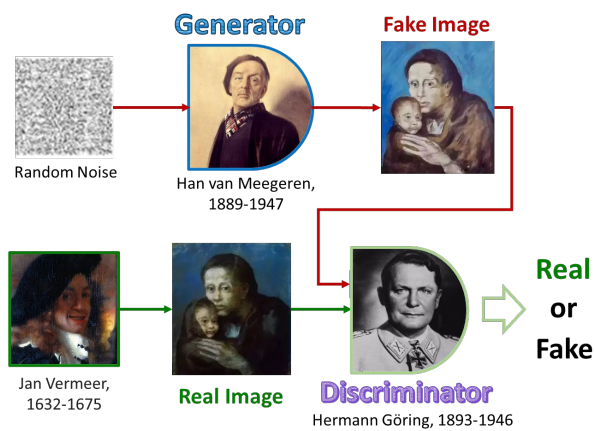

Figure 4: The architecture of GAN.

## 2.5. Super Resolution

Super-resolution is a typical computer vision task which aims at reconstructing a high-resolution image from a low-resolution image. Specifically, there is a demand for recovery of missing resolution information on each slice of paintings, which is considered an in-plane resolution problem [13]. Deep learning is a new breakthrough technology that is a branch of machine learning. Many existing deep learning studies have addressed various applications such as classification, detection, tracking, pattern recognition, image segmentation, and parsing. They have also demonstrated robust performance of deep learning compared to other machine learning tools. Deep learning-based single image SR methods have been recently introduced in computer vision [13]. Deep learning techniques greatly improve the performance of Super-resolution. The first super-resolution deep learning [14] consisted of three convolutional layers, and one fully connected layer. The capacity of convolutional neural network (CNN) expands with increasing depth and width, resulting in a significant improvement in super-resolution (as shown in Figure 5). Then, a multi-layer perceptron, where all layers are fully connected, is suitable for natural image denoising [15] and post-blurring denoising [16].

More closely related to our work, convolutional neural networks are applied to natural image denoising [17] and to remove noisy patterns [18]. These restoration problems are more or less denoising driven. Cui et al. [19] proposed to embed an autoencoder network in their super-resolution pipeline under the concept of the internal example-based method [20]. Deep learning models are not specifically designed as an end-to-end solution, as each layer of the cascade requires independent optimization of the self-similar search process and the auto-encoder. In contrast, our proposed model SANet (Self-Attention Networks) optimizes the end-to-end mapping. Also, our model introduces channel-wise attention [21], which can capture the feature importance during convolution processes. Channel-wise attention aims to model the relationships between different channels with different semantic concepts. By focusing on a part of the channels of the input feature and deactivating non-related concepts, the models can focus on the concepts of interest.

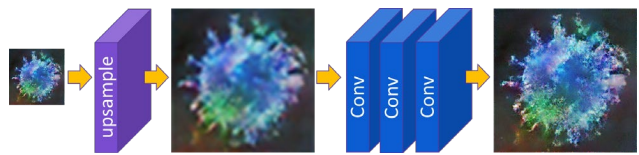

Figure 5: Super-resolution for SARS-CoV-2 painting.

## 2.6. Graph Neural Network

Graph Neural Network (GNN) have a wide range of applications in different tasks and domains. Each class of GNN has specialized general tasks including node classification, node representation learning, node clustering, graph classification, and graph partition [22]. Graph-based recommender systems treat items and users as nodes. By leveraging item-to-item, user-to-user, user-to-item relations, and content information, graph-based recommender systems are able to generate high quality recommendations. The key to recommender systems is to score the importance of items to users. As a result, it can be transformed into a link prediction problem. Ying et al. [23] propose a GNN-based graph auto-encoder to predict the missing link between users and items. Monty et al. [24] combines GNN and RNN to learn the underlying process that generates the known ratings.

In chemistry, researchers apply GNN to study the graph structure of molecules. In a molecular graph, atoms function as nodes and chemical bonds function as edges. Node classification, graph classification, and graph generation are three main tasks for molecular graphs of molecular fingerprints learning [25], molecular properties prediction [24], protein interfaces inference [26], and chemical compounds synthesizing [27]. Some scholars have initially explored the application of GNN to other problems, such as adversarial attacks prevention [28], electronic health records modeling [29], event detection [30], combinatorial optimization [31], program verification [32], program reasoning [33], and social influence prediction [34]. In this study, we use GNN to automatically learn the sequential relationship between pictures according to the style, texture, and color of the SARS-COV-2 paintings, resulting in the effect of gradual evolution between pictures.

## 3. Our Artwork: Medium & Permeation

Coronavirus continues to mutate, and mutant strains continue to emerge. The generation of mutant strains is the result of mutations occurring during the self-replication process of the virus. The virus will continue to replicate itself throughout its life. During this period, mutations will often occur, resulting in mutant strains. The more replication, the more infected people, the greater the probability of mutation and the greater the number of mutations. After SRAS-CoV-2 discovered in the end of 2019, the Alfa variant appeared in the spring of 2021, the Delta variant appeared in the summer, and then the Gamm variant appeared one after another. By the end of 2021, the Omicron variant appeared, and five different virus gene sequences have appeared. Therefore, we try to use Generative Adversarial Network, which allows the computer to draw the appearance of the virus by itself, after that through the Graph Neural Network, we make viruses randomly emerge color gradient changes to visualize the phenomenon of virus self-mutation.

### 3.1. Creative Motivation

The first industrial revolution prompted Impressionist painters to apply the scientific evidence of light, shadow, and color theory as the basis for their creative concepts. Now in the fourth industrial revolution, with the emergence of artificial intelligence, how can we apply the new technology to the expression of light, shadow and color in painting and echo the changes with the context of the nineteenth century Impressionism is our motivation for artistic creation.

Impressionists reflect the instant impression of nature according to the seven colors of red, orange, yellow, green, blue, indigo, and violet presented by the solar spectrum. In the 1870s, the French Impressionist Renoir often used pigments such as lead white, cadmium yellow, Napoli yellow, ochre, rich natural yellow, vermilion, marigold, Verona green, emerald-green, jade-green, cobalt blue, and ultramarine. Another Impressionist master, Monet, mainly used the following colors: lead white, chrome yellow, cadmium yellow, bright green, sapphire green, ultramarine, cobalt blue, alizarin red, and vermilion.

After analyzing the RGB values used in painting colors from the representative Impressionist painters: Renoir, Monet, Pissarro, and Degas, etc. through CBIR (Content-based image retrieval) image content search and retrieval Feature Extraction, as well as research on Impressionist color literature, we collected and sorted out Impressionist paintings data of the following features [35]: (1) The Feature Extraction of Dominant Color, (2) Adjacent Color Combination, (3) Color Structure Descriptor, (4) Color Layout Descriptor.

Next, we vectorized the above color information through machine learning, and formed the dimension of color through the vector. The serial value of the dimension reflects the main structure and layout of the color. Then we used the 300 virus illustrations collected around the world as the data base to train the computer to recognize the virus outline (pixel boundary), and then fed the generator and discriminator Conditional Generative Adversarial Network (cGAN) with conditional data at the same time.

The computer then learned to draw a virus with Impressionist style. In other words, the Impressionist "color code", which includes the numerical value of the main color pixel characteristics, color structure characteristics, and color layout characteristics of Impressionism, is hidden in the drawing. We also adopted a deep convolutional neural network to take a low-resolution image as input and output a high-resolution image. The following (Figure 6-9) are sample images of the virus generated by the machine learning with the cGAN. Compared with the famous Impressionist paintings, we will discover that whether it is the color distribution of the main tones, the color overlapping, and the color brightness, AI paintings are very much like Impressionist style.

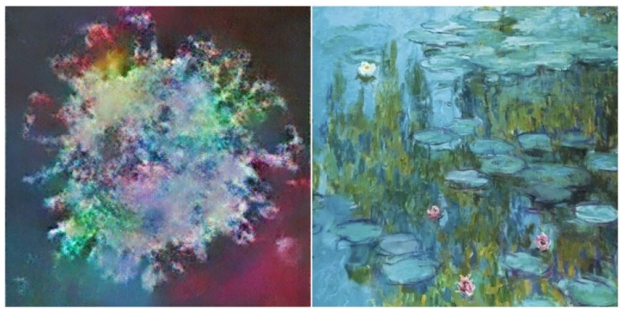

Figure 6: Right: Le Nymphéas by Claude Monet, 1915. Left: one of the randomly generated viruses from our work.

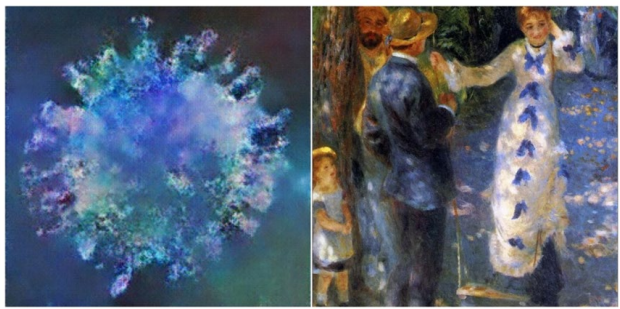

Figure 7: Right: La Balançoire by Pierre-Auguste Renoir, 1876. Left: one of the randomly generated viruses from our work.

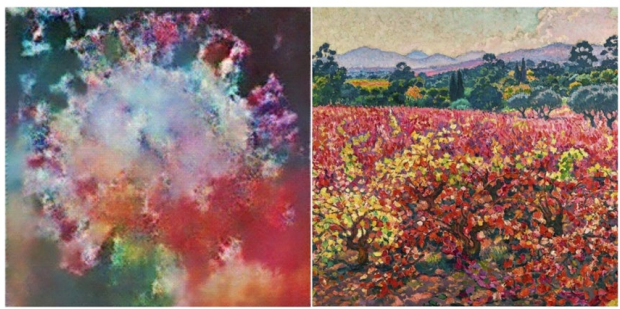

Figure 8: Right: La Vigne en Octobre by Theo van Rysselberghe, 1912. Left: one of the randomly generated viruses from our work.

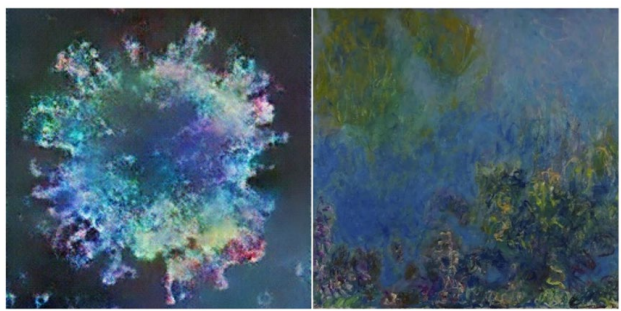

Figure 9: Right: Le Bassin aux Nymphéas by Claude Monet, 1917-1920. Left: one of the randomly generated viruses from our work.

## 3.2. Statement of Artwork

Virus particles smaller than 5μm can be suspended in the air temporarily, and they take a scattering route into the human body. When light encounters particles in the air, it produces a scattering state. The air is full of oxygen and nitrogen molecules. Their size is even smaller than the wavelength of the short wave, and the scattered light will be in the violet-blue-green range. When light in the blue light band is further scattered, red-orange-yellow colors appear.

Impressionism expressed the scientific significance of light in artistic creation. The light color reflected by the luminous flux through the medium in the air became the theoretical basis of the impressionist creation. Take Monet's paintings as an example, we see neither very well-defined shadows, nor prominent as well as flat-painted outlines. Monet's sensitivity to color is quite delicate. He experimented with the expression of color and light with many paintings of the same theme. He had long explored the performance effects of light, color and air. He often painted the same object multiple times at different moments and lighting, expressing the change from natural light and color.

We tried to use the impressionist style of painting to present the analogy that the diffuse paths of SARS-CoV-2 through the air are like the scattering phenomenon of light molecules caused by the air, an image of virions with no obvious outlines and shadows, but with overlapping colors.

## 4. SARS-COV-2 Paintings Generating Method

### 4.1. Self-Attention Generative Adversarial Network

An important part of art-generating algorithms is relating their creative process to the art that has been produced by human artists throughout time. We believe this is important because the human creative process utilizes prior experience and exposure to art. Human artists are continuously exposed to the work of other artists and have been exposed to a wide variety of art throughout their lifetime. What remains largely unknown is how human artists combine their knowledge of past art with their ability to create new forms. A theory is needed to model how to integrate exposure to art with the creation of art.

Martindale [36] proposed a theory based on psychology to explain new art creation. He hypothesizes that at any point in time, creative artists try to increase the arousal potential of their art in order to push against habituation. However, this increase must be minimal to avoid negative observer reactions. Martindale also hypothesized that style breaks happen as a way of increasing the arousal potential of art when artists exert other means within the roles of style. The approach proposed in this paper is inspired by Martindale's principle of least effort and his explanation of style breaks. In trying to explain the theory of artistic progress, we find that Martindale's theory is computationally feasible.

We propose an art-generating model that aims to generate artworks with increased levels of arousal potential in a constrained way without activating the aversion system and falling into the negative hedonic range. The art-generating model has a memory that encodes the art it has been exposed to, and can be continuously updated with the addition of new art. The art-generating uses this encoded memory in an indirect way, while generating new art with a restrained increase in arousal potential. The proposed art-generating model is realized by a model called Self-Attention Generative Adversarial Network (SAGAN).

GANs consists of two adversarial models: a generative model G and a discriminative model D. Generator captures the data distribution, and Discriminator estimates the probability that a sample came from the training data rather than Generator. Both Generator and Discriminator are trained simultaneously. As the formula shows, z is the input of random noise. We adjust parameters for Generator to minimize (log function of one minus model d) and adjust parameters for Discriminator to minimize (log function of model d), as if they are following the two-player min-max game with value function of model G and D.

$$\min_G \max_D V(D, G) = \mathbb{E}_x[log\, D(x|y, s)] + E_z[log(1 - D(G(z|y, s)))] \quad (1)$$

Conditional GANs are an extension of the GANs model. Like classical GANs, Conditional GANs also has two components. Generator and a Discriminator. They both receive some additional conditioning input information. As the formula shows, it has one more condition term y than classical GANs. This could be the class of the current image or some other property. The additional condition in our model is the country information of authors. For this addition, we add an additional input layer with values of one-hot-encoded image labels. Generator can learn the style and texture of various countries (as shown in Figure 10).

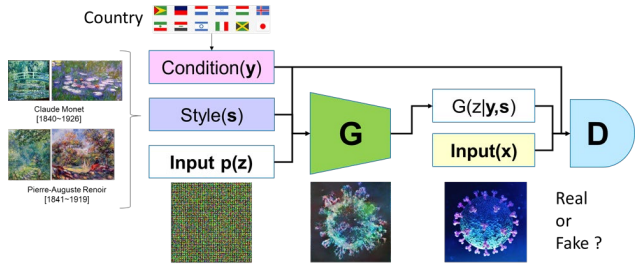

Figure 10: Our proposed model - SAGAN.

The generator architecture consists of two paths. First, the left hand side of architecture is called encoder path. This path consists of stack of convolutional layer and max pooling layer. This is used to capture the context of the image. Second, the right hand side of architecture is called decoder path. The path consists of transposed convolutional layers. This is used to expand the enable precise localization. For up-sampling, Transposed Convolutional layer is used. Parameters are in such a way that the height and width of the image will be doubled. In addition, we added self-attention block in the convolution process, as shown in Figure 11 by the red arrow.

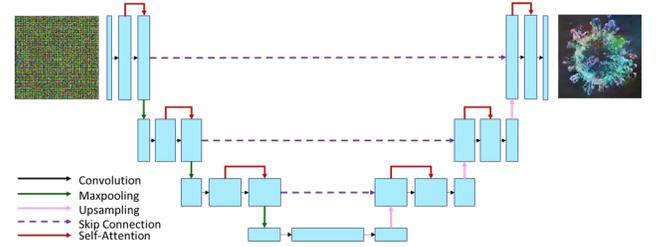

Figure 11: The architecture of Generator.

For each query location, self-attention block calculates the pairwise relationship between the query location and all locations to form an attention map, and then aggregates the features of all positions through the weighted sum with the weights defined by the attention map. Finally, the aggregated features are added to the features of each query position to form the output. The basic self-attention architecture is shown in the following formula:

$$m_i = \sum_{j=R(i)} \alpha(x_i, x_j) \odot \beta(x_j) \quad (2)$$

where $\odot$ circle is the Hadamard product, $i$ is the spatial index of feature vector $x_i$, which means the location in the feature map. $R(i)$ is the local footprint of the aggregation. The footprint $R(i)$ is a set of indices that specifies which feature vectors are aggregated to construct the new feature $m_j$. The function $\beta$ produces the feature vectors $\beta(x_j)$ that are aggregated by the adaptive weight vector function $\alpha(x_j, x_j)$. By attention mechanism, we can develop better methods for capturing the long-range dependencies of feature map (as shown in Figure 12).

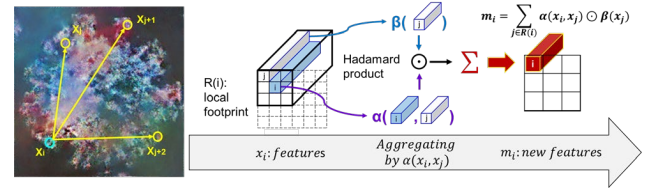

Figure 12: Self-Attention in Generator.

### 4.2. Super Resolution

Super-resolution is a long-studied technology, which aims to generate a high resolution visual-pleasing image from a low-resolution input. The aim of this section is to produce an high resolution SARS-COV-2 paintings, i.e., larger matrix size with extrapolated signals, from an original paintings of 256×256 pixels. We study how to apply super-resolution algorithms on large input SARS-COV-2 painting, which will be upsampled to at least 1K resolution (1000×1000). Considering a single low-resolution image, we first upscale it to the desired size using bicubic interpolation. Then, we try to recover the image to the high resolution by our proposed model Self-Attention-Super-Resolution Network (SASR-Net).

When the SARS-COV-2 original painting is upscaled, the image quality of the original painting slice will naturally decrease without compensating for the missing resolution information. Although the original paintings has inadequate spatial resolution for use in anatomical SARS-COV-2 paintings, the edges shown in the paintings are sufficiently sharp. Blurring should not be ignored in low-resolution images. In the downsampling method, bicubic interpolation results in a blurry image rather than pixelated images by calculating a weighted average of the nearest pixels. Therefore, we model the loss of pixel information and blurring caused by bicubic downsampling as Eq. (3):

$$Y = DS^f_{bicubic} X \qquad (3)$$

where $Y$ denotes a low resolution paintings corresponding to the original SARS-COV-2 paintings, $DS$ indicates the bicubic downsampling operator with a scaling factor $f$, and $X$ is a high resolution paintings corresponding to the enlarged paintings that we desire to obtain.

The proposed algorithm produces high resolution paintings from the original paintings $Y$ with a scaling factor $f$ and a parameter set $\Theta$. We define an outcome of our model as the symbol $Z$. The lose function for the bicubic downsampling is produced from Eq. (4), as follows:

$$L_D(Z) = Y - DS^f_{bicubic} Z \qquad (4)$$

The observation model is used to generate a training input dataset and the low-resolution paintings in our experiments. With Eq. (4), we can translate the given image super resolution problem of SARS-COV-2 paintings into an optimization problem:

$$\hat{X} = \left\{ \underset{z}{\mathrm{argmin}} \|L_D(Z)\|^2 : Z = F(Y; f; \Theta) \right\} \qquad (5)$$

where $\hat{X}$ indicates an estimated high-resolution painting, $Z$ is an outcome of SASR-Net, $F(Y; f; \Theta)$ denotes the proposed method as a function, and $\Theta$ is a parameter set for function. The parameter set $\Theta$ includes weights and biases of each layer in SASR-Net.

SASR-Net consists of convolution layers, attention block, and a deconvolution layer as illustrated in Fig. 13. Convolution layers and activation function layers are primary components of typical CNN. The other prime component of CNN is a pooling layer, also called a subsampling layer. The pooling layer chooses featured values from the image for progressive reduction of the number of parameters and the computational cost in the network, but it causes loss of input image information. In order to keep the feature values, the proposed method excluded the pooling layer in its architecture and designed SASR-Net with the deconvolution layer to upsample the low-resolution SARS-COV-2 paintings. We also introduce channel-wise attention in our model, which can capture the feature importance during convolution processes. By focusing on a part of the channels of the input feature and deactivating non-related concepts, the models can focus on the concepts of interest.

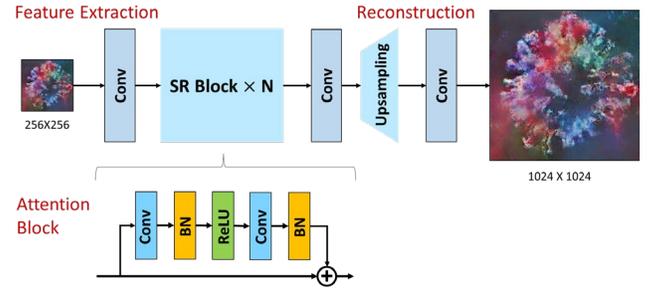

Figure 13: The architecture of SASR-Net.

### 4.3. Gradual Change by Graph Neural Network

In this section, we have developed an extended work: Gradual Change. We use Graph Neural Network technology to present 196 paintings of the new coronavirus created by A.I. to the audience one by one in a gradual manner. It symbolizes that the current new coronavirus will continue to evolve, and the speed of vaccine development cannot keep up with the speed of virus mutation. This work symbolizes that SARS-CoV-2 has brought challenges that the world has never had before.

Graph Neural Network is a framework for unsupervised learning on graph-structured data based on the variational auto-encoder. This model makes use of latent variables and is capable of learning interpretable latent representations for undirected graphs. A graph simply consists of nodes and connections between these nodes, which we called edge. In this paper, the node represents one SARS-CoV-2 painting, and the edge represents relationship between the nodes. The information about these connections in graph can be represented in an adjacency matrix. The elements of the matrix indicate connected nodes with a 1 and disconnected nodes with a 0 (as shown in Figure 14).

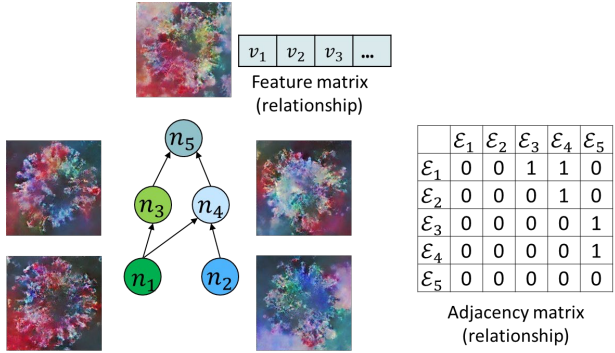

Figure 14: The left side of this figure is a Graph structure. The feature matrix V of each node comes from the feature extraction of the SARS-CoV-2 painting. The right side of this figure is an adjacency matrix, and each E_i represents the relationship with its neighbor.

The feature matrix $\mathcal{V}$ of each node comes from the feature extraction of the SARS-CoV-2 painting. The right side of this figure is an adjacency matrix, and each $\mathcal{E}_i$ represents the relationship with its neighbor. The graph model can be represented by the following formula:

$$\mathcal{G} = (\mathcal{V}, \mathcal{E}) \qquad (6)$$

where $\mathcal{V}$ is the feature matrix, and $\mathcal{E}$ is the adjacency matrix. The goal of this model is then to learn a function of features on a graph $\mathcal{G}$. A feature description $v_i$ for every node $i$ summarized in a $N \times D$ feature matrix $\mathcal{V}$. $N$ is the number of nodes; $D$ is number of input features. A representative description of the graph structure in matrix form, which is typically in the form of an adjacency matrix $\mathcal{E}$. Then, the model produces a node-level output $z$. It is presented with an $N \times F$ feature matrix, where $F$ is the number of output features per node. Graph-level outputs can be modeled by introducing pooling operation.

Every neural network layer can then be written as a non-linear function. Let's consider the following form of a layer-wise propagation rule:

$$f(H^{(l)}, A) = \sigma(AH^{(l)}W^{(l)}) \qquad (7)$$

where $W^{(l)}$ is a weight matrix for the $l$-th neural network layer and $\sigma(\cdot)$ is a non-linear activation function of rectified linear unit (ReLU). $H^{(l)}$ multiplication with $A$ means that, for every node, we sum up all the feature vectors of all neighboring nodes. Then, Graph Convolutional layer-wise propagation rule as the following formula:

$$h_i^{(l+1)} = \sigma(\sum_{j \in N(i)} \frac{1}{c_{ij}} W^{(l)} h_j^{(l)}) \qquad (8)$$

where $j$ indexes the neighboring nodes of $i$. $c_{ij}$ is a normalization constant for the edge ($v_i$, $v_j$) which originates from using the symmetrically normalized adjacency matrix in our GNN model. This propagation rule can be interpreted as a differentiable and parameterized variant of the hash function. Furthermore, we choose ReLU as nonlinearity function and initialize random weight matrix. This update method becomes stable during model training. After mean aggregation of all neighbor nodes and activation by ReLU, the features of each node $j$ will be update into the node $i$ (as shown in Figure 15).

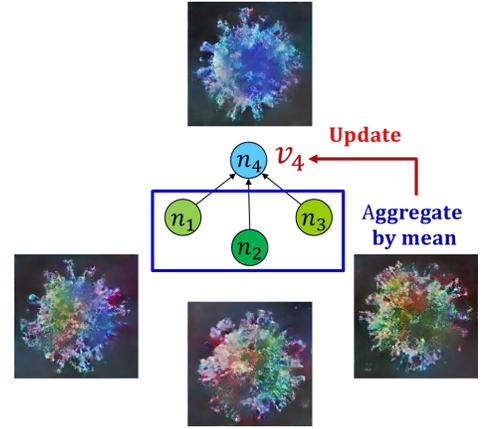

Figure 15: Aggregation and feature update from neighboring nodes.

## 5. Generative Results

Our works generated 196 paintings through the Self-Attention Generative Adversarial Network after learning the creation of SARS-COV-2 art works by artists from all over the world. We trained the generative model and let the computer randomly synthesize it. We adjusted various parameters dur-ing the training process making the results such as Dou Jia, Monet, Renoir, their style paintings. Then, we use Self-Attention Super Resolution Network to obtain a higher resolution painting. Through SASR-Net, we can build a deeper network for feature extraction without the vanishing gradient problems. The final artwork of SARS-COV-2 paintings produced by artificial intelligence is shown in Figure 16.

Finally, we have developed an extended work: Gradual Change. We use Graph Neural Network to present 196 painting of the new coronavirus created by SAGAN to the audience one by one in a gradual manner. It symbolizes that the current new coronavirus will continue to evolve, and the speed of vaccine development cannot keep up with the speed of virus mutation. This work symbolizes that SARS-CoV-2 has brought challenges that the world has never had before. Gradual Change is presented in video, please refer as: *https://www.youtube.com/embed/vpkR4jU1aec*.

## 6. Conclusion

While deepening the world's impression of the virus catastrophe, the work also has the so-called artistic aesthetics. People are afraid of viruses that cannot be eliminated, but we hope that the creation itself can free the viewer from fear, so that they can approach the work.

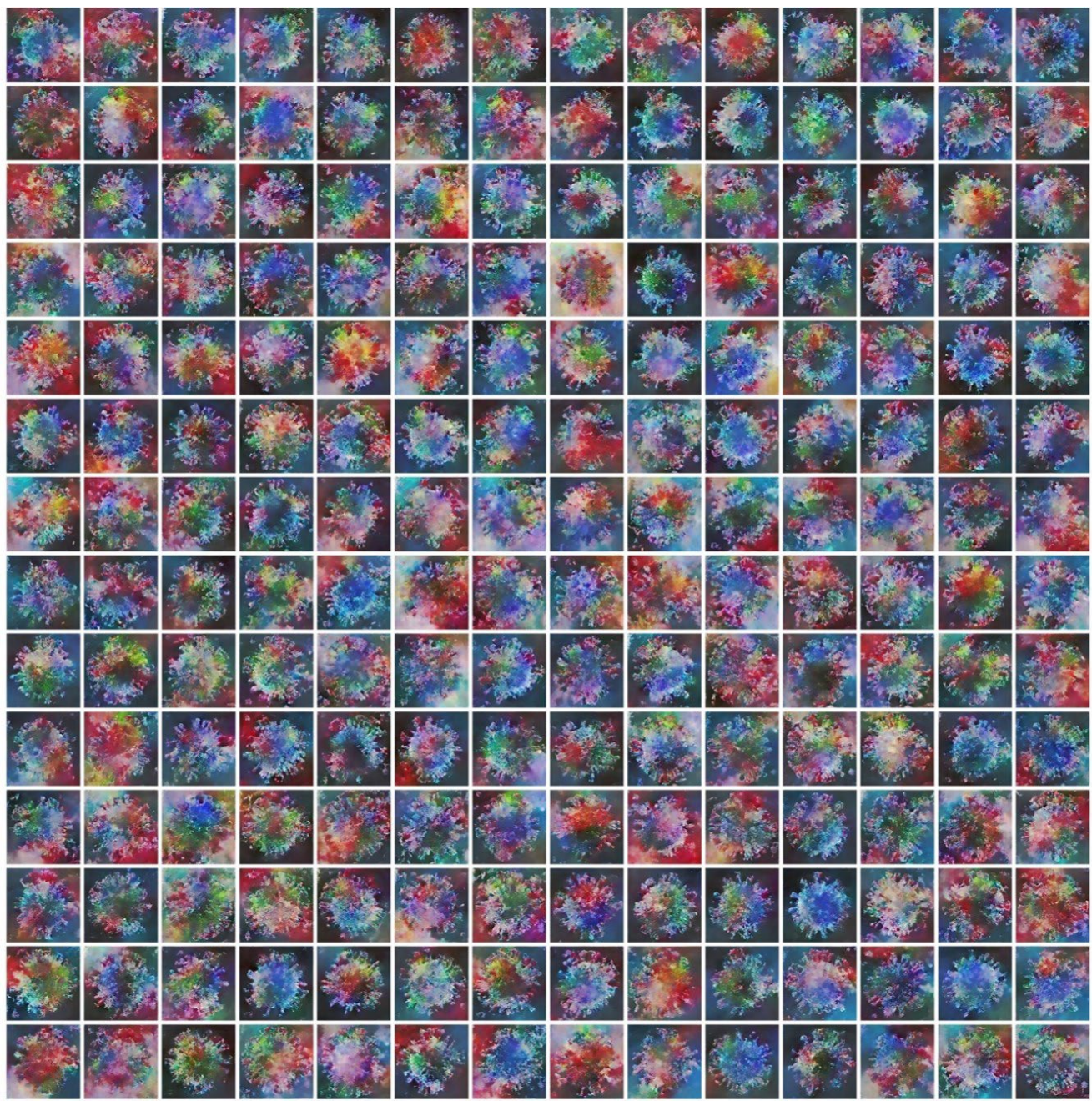

Figure 16: Medium. Permeation is composed of 196 randomly generated SARS-COV-2 pictures arranged in a 14 by 14 matrix to form a large-scale painting. We train SAGAN to draw the novel coronavirus in Impressionist style with computerized data analysis through the main tones, the way of color layout, and the way of color stacking in the paintings of the Impressionists.

Bright colors can make the viewer willing to stop for not just a glance. However, bright colors are a warning signal from nature. The closer the viewer gets to the colorful dynamic images that change in real time, the closer he is to the dangerous virus and metaphorically becomes a member of the transmission chain. Not only the results of our creation, but even the process of creation, reflect the symbiosis of viruses and people, and the symbiosis of impressionism and artificial intelligence. And through the fourth industrial revolution of mankind, using generators and discriminators in the principle of generative adversarial networks, artificial intelligence allows Impressionism and viruses to meet across space and time, depend on each other, and grow together.

In the future, we plan to systematically convert images of other schools of painting into data analysis based on the color styles and techniques through artificial intelligence. We will continue with our work of sorting out the genealogy of the digital codes hidden in the paintings of different schools. Hopefully, this work will help to build the scientific data system for chromatic study, which will enrich the field of Western art history.

## References

[1] Alice Rawsthron, "Alissa Eckert on designing the spiky blob Covid-19 medical illustration", *Wallpaper official website*, Sep 26, 2020, *https://www.wallpaper.com/design/design-emergency-alissa-eckert-designs-covid-19-illustration*

[2] David S. Goodsell, "Molecular landscapes", *Protein Data Bank*, *https://pdb101.rcsb.org/sci-art/goodsell-gallery*

[3] A. Lilja and J. McGhee, "Soap vs COVID-19: A 3D-visualisation gamifies the power of simple hygiene," *University of New South Wales website*, July 28, 2020, *https://www.unsw.edu.au/news/2020/07/soap-vs-covid-19--a-3d-visualisation-gamifies-the-power-of-simpl*

[4] Richard Allen Shiff, "Impressionist criticism , impressionist color, and cezanne", *Yale University ProQuest Dissertations Publishing*, 1973.7329249.

[5] Laura Anne Kalba, "Color in the age of impressionism: commerce, technology, and art", *University Park: Penn State University Press*, 2017.

[6] Philip Galanter, "What is generative art? Complexity theory as a context for art theory", *Generative Art Conference*, pp. 225–245, 2003.

[7] Marc Lee, "Corona TV Bot", *Personal website, accessed Mar 20, 2020*, *https://marclee.io/en/new-media-art-reflects-thecoronavirus-pandemic/*

[8] Marc Lee, "Corona TV Bot", *Annka Kultys Gallery website*, Apr. 20, 2020, *http://www.annkakultys.com/online/stay-at-home/stay-at-home-marc-lee/*

[9] Alain Lioret, "Quantum generative art department of arts et Technologies de l'Image", *Generative Art Conference*, 2021.

[10] Steve DiPaola and Liane Gabora, "Incorporating characteristics of human creativity into an evolutionary art algorithm", *Genetic Programming and Evolvable Machines*, pp. 97–110, 2009.

[11] Jeanine Graf and Wolfgang Banzhaf, "Interactive evolution of images", *In Evolutionary Programming*, pp. 53–65, 1995.

[12] Ian Goodfellow, et al., "Generative adversarial nets", *Neural Information Processing Systems (NeurIPS)*, 2014.

[13] Eric Van Reeth, Ivan W. K. Tham, Cher Heng Tan, and Chueh Loo Poh, "Super-resolution in magnetic resonance imaging: A review", *Concepts in Magnetic Resonance*, pp. 306-325, Nov. 2012.

[14] Chao Dong, Chen Change Loy, Kaiming He, and Xiaoou Tang, "Learning a deep convolutional network for image super-resolution", *European Conference on Computer Vision (ECCV)*, 2014.

[15] Harold C. Burger, Christian J. Schuler, and Stefan Harmeling, "Image denoising: Can plain neural networks compete with BM3D", *Conference on Computer Vision and Pattern Recognition (CVPR)*, 2012.

[16] Christian J. Schuler, Harold Christopher Burger, Stefan Harmeling, and Bernhard Schölkopf, "A machine learning approach for non-blind image deconvolution", *Conference on Computer Vision and Pattern Recognition (CVPR)*, 2013.

[17] Viren Jain and Sebastian Seung, "Natural image denoising with convolutional networks", *Neural Information Processing Systems (NeurIPS)*, 2008.

[18] David Eigen, Dilip Krishnan, and Rob Fergus, "Restoring an image taken through a window covered with dirt or rain", *IEEE Inter-national Conference on Computer Vision (ICCV)*, 2013.

[19] Zhen Cui, Hong Chang, Shiguang Shan, Bineng Zhong, and Xilin Chen, "Deep network cascade for image super-resolution", *European Conference on Computer Vision (ECCV)*, 2014.

[20] Daniel Glasner, Shai Bagon, and Michal Irani, "Super-resolution from a single image", *IEEE International Conference on Computer Vision (ICCV)*, 2019.

[21] Jie Hu, Li Shen, Samuel Albanie, Gang Sun, and Enhua Wu, "Squeeze-and-Excitation Networks", *Conference on Computer Vision and Pattern Recognition (CVPR)*, 2018.

[22] Guimin Dong, Mingyue Tang, Zhiyuan Wang, Jiechao Gao, Sikun Guo, Lihua Cai, Robert Gutierrez, Bradford Campbell, Laura E. Barnes, and Mehdi Boukhechba, "Graph neural networks in IoT: A survey", *arXiv:2203.15935*, 2022.

[23] Rex Ying, Ruining He, Kaifeng Chen, Pong Eksombatchai, William L. Hamilton, and Jure Leskovec, "Graph convolutional neural networks for web-scale recommender systems", *International Conference on Knowledge Discovery and Data Mining*, pp. 974–983, 2018.

[24] Federico Monti, Michael M. Bronstein, and Xavier Bresson, "Geometric matrix completion with recurrent multi-graph neural networks", *Neural Information Processing Systems (NeurIPS)*, pp. 3697–3707, 2017.

[25] Steven Kearnes, Kevin McCloskey, Marc Berndl, Vijay Pande, and Patrick Riley, "Molecular graph convolutions: moving beyond fingerprints," *Journal of computer-aided molecular design*, vol. 30, no. 8, pp. 595–608, 2016.

[26] Justin Gilmer, Samuel S. Schoenholz, Patrick F. Riley, Oriol Vinyals, and George E. Dahl, "Neural message passing for

quantum chemistry", *International Conference on Machine Learning (ICML)*, pp. 1263–1272, 2017.

[27] Jiaxuan You, Bowen Liu, Rex Ying, Vijay Pande, and Jure Leskovec, "Graph convolutional policy network for goal-directed molecular graph generation", *Neural Information Processing Systems (NeurIPS)*, 2018.

[28] Daniel Zügner, Amir Akbarnejad, and Stephan Günnemann, "Adversarial attacks on neural networks for graph data", International *Conference on Knowledge Discovery and Data Mining*, pp. 2847–2856, 2018.

[29] Edward Choi, Mohammad Taha Bahadori, Le Song, Walter F. Stewart, and Jimeng Sun, "Gram: graph-based attention model for healthcare representation learning", *International Conference on Knowledge Discovery and Data Mining*, pp. 787–795, 2017.

[30] Thien Nguyen and Ralph Grishman, "Graph convolutional networks with argument-aware pooling for event detection", *Association for the Advancement of Artificial Intelligence (AAAI)*, pp. 5900–5907, 2018.

[31] Zhuwen Li, Qifeng Chen, and Vladlen Koltun, "Combinatorial optimization with graph convolutional networks and guided tree search", *Neural Information Processing Systems (NeurIPS)*, pp. 536–545, 2018.

[32] Yujia Li, Daniel Tarlow, Marc Brockschmidt, and Richard Zemel, "Gated graph sequence neural networks", *International Conference on Learning Representations (ICLR)*, 2015.

[33] Miltiadis Allamanis, Marc Brockschmidt, and Mahmoud Khademi, "Learning to represent programs with graphs", *International Conference on Learning Representations (ICLR)*, 2017.

[34] Jiezhong Qiu, Jian Tang, Hao Ma, Yuxiao Dong, Kuansan Wang, and Jie Tang, "DeepInf: Social influence prediction with deep learning", *International Conference on Knowledge Discovery & Data Mining*, pp. 2110–2119, 2018.

[35] Man-Kwan Shan, "Discovering color styles from fine art images of impressionism", *International Journal of Computer Science and Security (IJCSS)*, pp. 314-324, Oct. 2009.

[36] Colin Martindale, "The clockwork muse: The predictability of artistic change", *Journal of Aesthetics and Art Criticism*, pp. 171-173, 1992.